\documentclass[lettersize,journal]{IEEEtran}
\usepackage{amsmath,amsfonts}
\usepackage{algorithmic}
\usepackage{array}
\usepackage[caption=false,font=normalsize,labelfont=sf,textfont=sf]{subfig}
\usepackage{textcomp}
\usepackage{stfloats}
\usepackage{url}
\usepackage{svg}
\usepackage{hhline}
\usepackage{bm}
\usepackage{booktabs}
\usepackage{multirow}
\usepackage{verbatim}
\usepackage{graphicx}
\def\BibTeX{{\rm B\kern-.05em{\sc i\kern-.025em b}\kern-.08em
    T\kern-.1667em\lower.7ex\hbox{E}\kern-.125emX}}
\usepackage{balance}
\begin{document}
\title{HyperDefect-YOLO: Enhance YOLO with HyperGraph Computation for Industrial Defect Detection}
\author{Zuo Zuo, Jiahao Dong, Yue Gao, \IEEEmembership{Senior Member, IEEE,} and Zongze Wu, \IEEEmembership{Member, IEEE}
\thanks{Manuscript created November, 2024. (Corresponding author: Zongze Wu.)

Zuo Zuo is with National Key Laboratory of Human-Machine Hybrid Augmented Intelligence, National Engineering Research Center for Visual Information and Applications, and Institute of Artificial Intelligence and Robotics, Xi'an Jiaotong University, Xi’an 710049, China (E-mail: Nostalgiaz@stu.xjtu.edu.cn).

Jiahao Dong is with Guangdong Laboratory of Artificial Intelligence and Digital Economy (SZ) (E-mail: dongjiahao2023@email.szu.edu.cn).

Yue Gao are with the School of Software, BNRist, THUIBCS, BLBCI, Tsinghua University, Beijing 100084, China. (E-mail: gaoyue@tsinghua.edu.cn).

Zongze Wu is with College of Mechatronics and Control Engineering, Shenzhen University, Shenzhen, China and Institute of Artificial Intelligence and Robotics, Xi'an Jiaotong University, Xi’an 710049, China (E-mail: zzwu@szu.edu.cn).
}}


\markboth{Journal of \LaTeX\ Class Files,~Vol.~18, No.~9, September~2020}%
{How to Use the IEEEtran \LaTeX \ Templates}

\maketitle

\begin{abstract}
In the manufacturing industry, defect detection is an essential but challenging task aiming to detect defects generated in the process of production.  Though traditional YOLO models presents a good performance in defect detection, they still have limitations in capturing  high-order feature interrelationships, which hurdles defect detection in the complex scenarios and across the scales. To this end, we introduce  hypergraph computation into YOLO framework, dubbed HyperDefect-YOLO (HD-YOLO), to improve representative ability and semantic exploitation.  HD-YOLO consists of Defect Aware Module (DAM) and Mixed Graph Network (MGNet) in the backbone, which specialize for perception and extraction of defect features. To effectively aggregate multi-scale features, we propose HyperGraph Aggregation Network (HGANet) which combines hypergraph and attention mechanism to aggregate multi-scale features. Cross-Scale Fusion (CSF) is proposed to adaptively fuse and handle features instead of simple concatenation and convolution. Finally, we propose Semantic Aware Module (SAM) in the neck to enhance semantic exploitation for accurately localizing defects with different sizes in the disturbed background. HD-YOLO undergoes rigorous evaluation on public HRIPCB and NEU-DET datasets with significant improvements compared to state-of-the-art methods. We also evaluate HD-YOLO on self-built MINILED dataset collected in real industrial scenarios to demonstrate the effectiveness of the proposed method. The source codes are at https://github.com/Jay-zzcoder/HD-YOLO.
\end{abstract}


\section{Introduction}
\IEEEPARstart{I}{ndustrial} defect detection aiming to identify and localize various defects is an indispensable link in industrial manufacturing chain~\cite{pmi}, which guarantees the quality and safety of products and is used for fault analysis~\cite{moninet} and production optimization. Deep learning methods~\cite{cablock,egf} have played a pivot role in defect detection recently. Deep defect detection methods can be classified as three categories: object-detection-based methods~\cite{taanet}, segmentation-based methods~\cite{clipfsac} and unsupervised defect detection~\cite{rfa}.  In this paper, we focus on object-detection-based defect detection including one-stage~\cite{yolov8}, two-stage~\cite{fasterrcnn} and transformer-based methods~\cite{rtdetr}.

\begin{figure}
  \centering
  \centerline{\includegraphics[width=\linewidth]{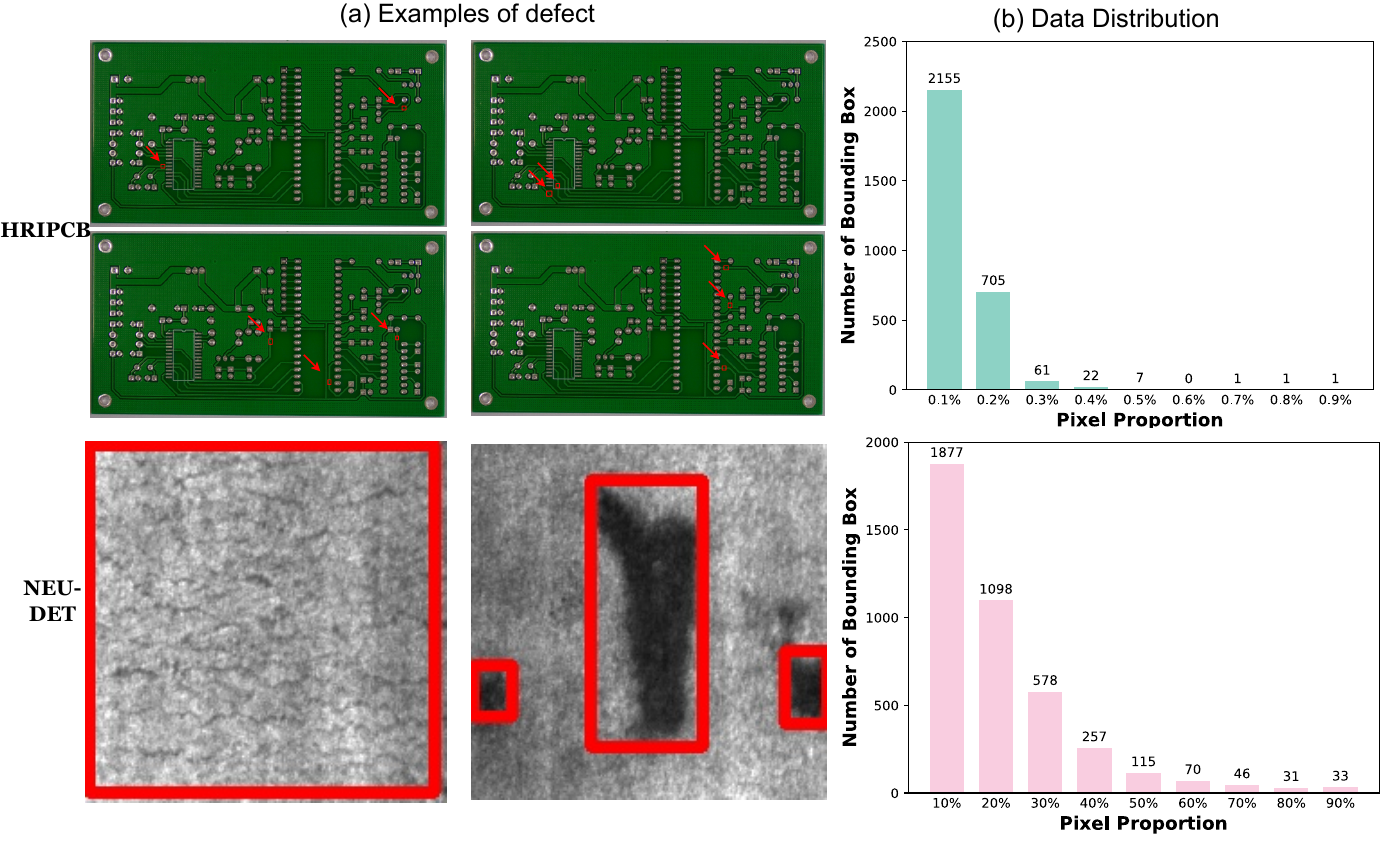}}
  \caption{Examples and data distribution of different datasets.}
  \label{datadis}
\end{figure}

You Only Look Once (YOLO) series is widely used in industrial defect detection. YOLO, as one-stage object detection methods, show strong detection ability, fast speed and low memory cost, achieving satisfactory performance in PCB~\cite{yolohmc}, transmission line insulator~\cite{adin}, optical lens~\cite{wgso}, etc. YOLO-HMC~\cite{yolohmc} based on YOLOv5 framework introduces the HorNet into backbone and proposes content-aware reassembly of features~\cite{carafe} as up-sampling layer for tiny-sized PCB defect detection. However, there are more than just tiny defects in real-world industrial scenarios. YOLO-HMC can not be effectively adapted for defects with different sizes. In \cite{drfa}, an anchor-free detector with dynamic receptive field assignment (DRFA)
and task alignment is proposed to extract complex defect features and dynamically adjust receptive field to detect diverse defects. While it is limited to extract high-order feature interrelationships and circumvent interference of complex background.
MCI-GLA Plug-In augmenting the learning capability of networks is proposed in~\cite{mcigla} to improve perception of global context information and local spatial details, which is used for transmission line insulator
defect detection. Defects on insulator are relatively obvious and background is simple which make the insulator defect detection easier.

In general, the difficulties which impede defect detection are summarized as follows:
\par 1) The sizes of defects change with industrial scenarios and types of products as shown in Fig. \ref{datadis}(a) which requires defect detection methods to adapt multi-scale defect features. It can be seen that defects in PCB are tiny shown in the first row while in strip steel the defects are sometimes very large shown in the second row. Statistics about size of defects are also given in Fig. \ref{datadis}. As can be seen, the proportion of defect pixel are less than 1\% and most are 0.1\% in HRIPCB~\cite{hripcb}, which means that defects in HRIPCB are so tiny. While in NEU-DET~\cite{neudet}, the proportion of defect pixel are larger than 10\% and some defects even occupy 90\% pixels. The size of defects in HRIPCB is far smaller than that in NEU-DET. 
\par 2) Complex background and unconspicuous defects pose a dilemma to defect detection. As shown in Fig. \ref{datadis}, the background of industrial images is complex, which interferes with extraction and process of defect features. Besides, unexpected appearance of unconspicuous defects which are hard to detect further makes defect detection more challenging.
\par 3) Defect features which are out of distribution break normal feature interrelationships. Therefore, exploration and extraction of  high-order feature interrelationships are vital in defect detection task. However, CNN was designed for low-order feature processing. We introduce hypergraph computation to compensate limitations of CNN.

\begin{figure}
  \centering
  \centerline{\includegraphics[width=\linewidth]{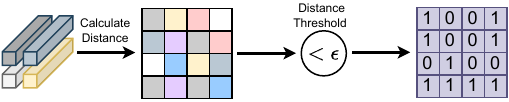}}
  \caption{The process of  hypergraph construction.}
  \label{hgc}
\end{figure}

To holistically address the issues mentioned above, we propose HyperDefect-YOLO (HD-YOLO) based on YOLO framework. To weaken unconcerned background features and concentrate on more defect features, we design Defect Aware
Module (DAM) in which Defect Aware Network (DAN) generates attention maps to enhance perception and extraction of defect features. Mixed Graph Network (MGNet) including hypergraph computation is for capturing feature interrelationships and downsampling in backbone.  Hypergraph computation can exploit high-order feature interrelationships and perceive out-of-distribution features that is defect features. Like Hyper-YOLO, we propose HyperGraph Aggregation Network (HGANet) as a bridge between backbone and neck which combines hypergraph and attention mechanism to aggregate multi-scale features in backbone. In HGANet, we propose an intriguing attention mechanism termed as Distance-Based Attention (DBA), in which we construct attention map based on distance between visual features. In neck, we design Cross-Scale Fusion (CSF) instead of inflexible concatenation to adaptively fuse features in different scales without loss of information. Neck is an essential part in detection neural networks which makes semantic feature transition to head for detection results. Consequently we propose Semantic Aware Module (SAM) in the neck to enhance semantic exploitation for accurately localizing defects with different sizes in the disturbed background. Our contributions can be summarized as:
\par 1) We propose the HyperDefect-YOLO (HD-YOLO) framework which can be used for enhancing defects feature perception and capturing of high-order feature correlations. It can be applied in different industrial scenarios for different products.

\par 2) To focus on defect features and eliminate influence of complex background, we propose DAM and MGNet in backbone and HGANet to aggregate multi-scale features to bridge backbone and neck, facilitating feature extraction and exploration. We also propose an inspiring distance-based attention to focus on key defect features.

\par 3) Due to various sizes of defects, our detection network needs to handle different-scale defect features. Therefore, we propose CSF to adaptively fuse and aggregate multi-scale features. Additionally, we incorporate SAM into neck to further enhance and enrich semantic features.

\par 4) We validate the effectiveness of HD-YOLO on PCB and steel plates from HRIPCB and NEU-DET, which are two totally different industrial datasets. Besides, we apply the proposed HD-YOLO to our self-built dataset: MINILED dataset. HD-YOLO achieves impressive defect detection performance on these datasets.

It is worth noting that although we build HD-YOLO based on YOLO framework, we redesign nearly all the modules in YOLO series framework.

The remainder of this article is organized as follows. The object detection, defect detection methods and hypergraph neural network are briefly reviewed in Section II and the proposed HD-YOLO model is introduced in detail in Section III. In Section IV, the effectiveness of HD-YOLO is verified on the public dataset and compared with the performance of other methods. Finally, some conclusions are drawn.

\begin{figure*}
\centering
\includegraphics[width=\textwidth,]{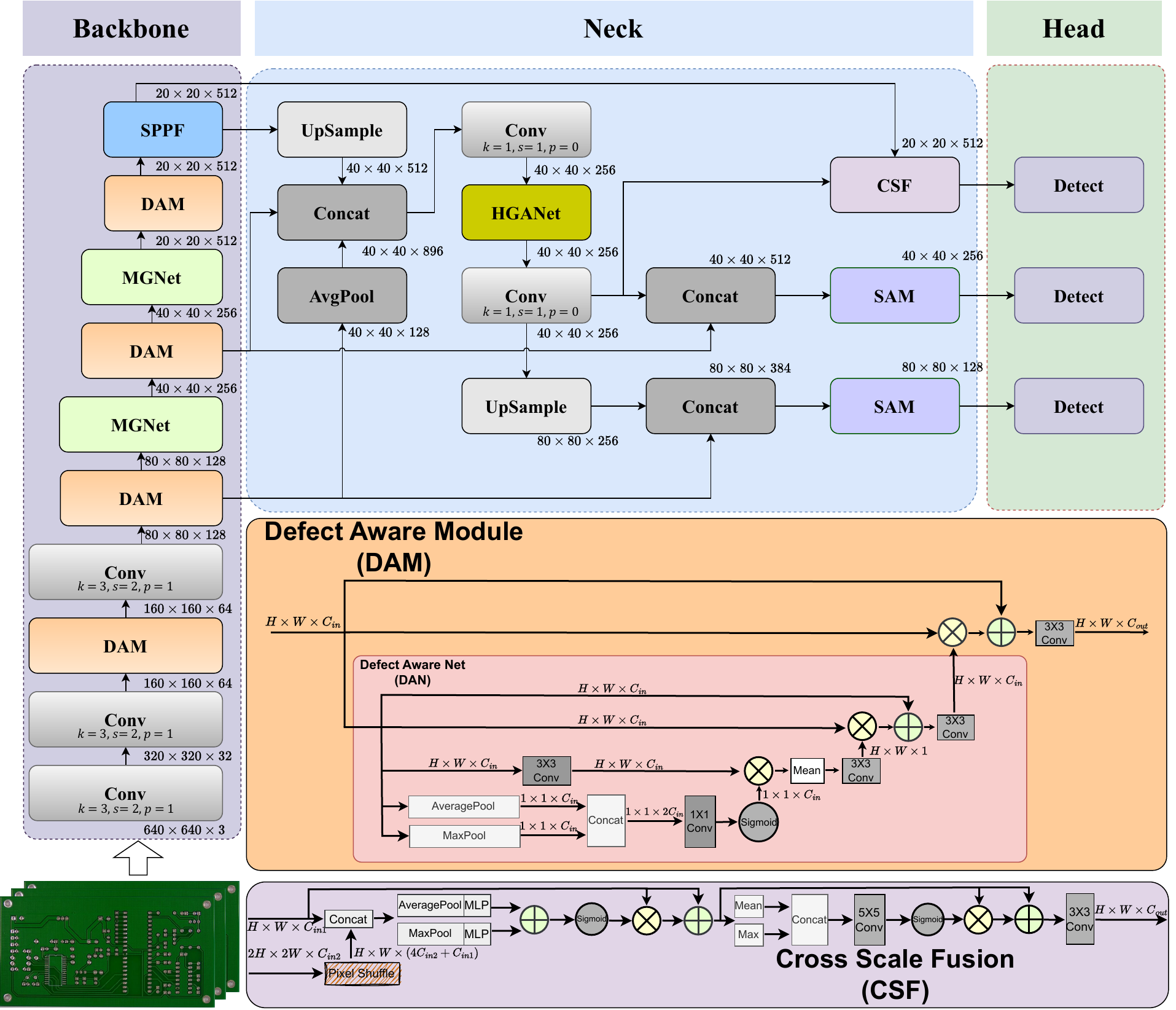}\\
\caption{Framework of the proposed HD-YOLO. }
\label{framework}
\end{figure*}

\section{RELATED WORK}

\subsection{Deep learning based defect detection}
Traditional methods are widely used in defect detection over the past. Development of deep learning significantly advance defect detection which avoids handcrafted features and leads to better performance~\cite{mfg}.  Unpredictable and diverse sizes and shapes of industrial defects pose a real-world dilemma to defect detection. Moreover, complex background seriously impede defect localization. SDDF-Net~\cite{sddfnet} incorporates spatial deformable convolution in backbone to extract complex features of defects effectively. PCB-YOLO~\cite{pcbyolo} introduces the attention mechanism to focus on defects and mitigate background interference in PCB defect detection. Zuo et al.~\cite{drfa} propose an efficient anchor-free detector where dynamic receptive field assignment (DRFA) and task alignment are introduced.
Besides, two-stage CANet~\cite{canet} proposes CAblock and LaplacianFPN for effective perception and exploitation of small defect features. Two-stage methods are often time-consuming and memory-consuming which is not friendly for real-world deployment. However, defect detection methods mentioned above can not fully exploit semantic visual features and capture high-order feature correlations which is vital in challenging defect detection.

\subsection{Hypergraph learning methods}
Different from graph-based convolutional neural networks, the hypergraph neural network performs well in dealing with complex data correlation and multi-modal data representation. In hypergraph, the hyperedge connects more than two vertices which provides a more flexible way to represent and handle complex relationships. Hypergraph Neural Networks (HGNN)~\cite{HGNN} where hyperedge convolution operation is proposed encodes high-order data correlation in a hypergraph structure and General Hypergraph Neural Networks (HGNN+)~\cite{HGNNP} introduces a general high-order multi-modal/multi-type data correlation modeling framework. The hypergraph learning has been broadly used in diverse tasks due to its strong modeling ability, such as social network analysis~\cite{lbsn2}, drug-target interaction modeling~\cite{hf}, brain network analysis~\cite{mhl}, etc. Besides, regarding the multi-modal data, HAN constructs symbolic graphs, hypergraphs and co-attention maps to extract joint representation of different modalities. Recently, HyperYOLO~\cite{hyperyolo} pioneers an object detection method based on hypergraph learning in which hypergraph convolution is proposed to capture the complex, high-order feature interrelationships. Inspired by Hyper-YOLO, we embed hypergraph computation in our proposed HD-YOLO framework to explore sophisticated feature interrelationships in industrial images by integrating the intricate relational information modeled by the hypergraph.

\subsection{YOLO Series Object Detectors}
YOLO series as one-stage object detection framework is widely used in object detection and they are optimized continually over the past years. 
The preliminary YOLO-v1 is firstly proposed in work~\cite{yolo}. It regards object detection as a regression problem. In the following, YOLO-v2~\cite{yolov2}, YOLO-v3 and YOLO-v4 are proposed sequentially by incorporating well-designed tricks and visual modules in YOLO framework. YOLO-v5 is so practical that it is used in many scenarios for real-world application. YOLOX~\cite{yolox} adopts anchor-free detection strategy to simplify anchor-based detection methods. YOLO-v6~\cite{yolov6} introduces RepVGG~\cite{repvgg} as backbone and adopts TOOD~\cite{tood} for label assignment. YOLO-v7 use E-ELAN backbone and a series of bag-of-freebies for optimization. In YOLO-v8~\cite{yolov8}, CIoU and DFL loss are used and anchor-free detection strategy remains.  In recent one year, YOLO-v9~\cite{yolov9} with programmable gradient information (PGI) and the Generalized Efficient Layer Aggregation Network (GELAN), YOLO-v10~\cite{yolov10} with dual assignments strategy and efficiency-accuracy driven model design strategy and YOLO-v11~\cite{yolov11} with updated model architecture are proposed rapidly. Hyper-YOLO breaks new ground and combines hypergraph computation and YOLO framework to improve the learning and integration of hierarchical features. Based on Hyper-YOLO and YOLO series framework, we present HD-YOLO specializing for industrial defect detection which pushes the limits of defect detection performance.

\begin{figure}
  \centering
  \centerline{\includegraphics[width=0.7\linewidth]{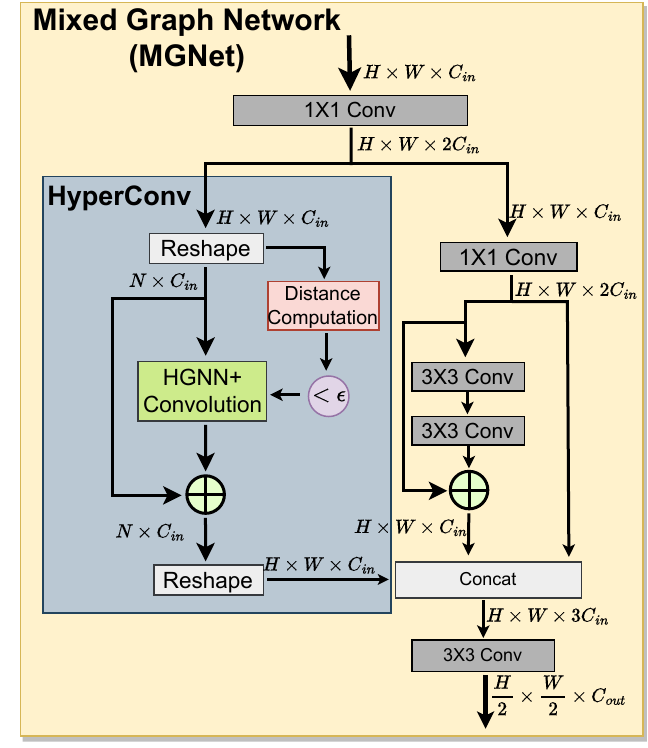}}
  \caption{Illustration of the proposed Mixed Graph Network (MGNet).}
  \label{mgnet}
\end{figure}

\section{METHODOLOGY}
In this section, we will delve into the framework of HD-YOLO and detail the proposed basic blocks (DAM and MGNet in backbone), HGANet, and (CSF, SAM) in neck.

\subsection{Preliminaries}
One of the highlights in HD-YOLO is combination of hypergraph convolution which is proposed in HGNN+~\cite{HGNNP} and is combined with YOLO in Hyper-YOLO. Hypergraph convolution regard visual features as vertex. We introduce hypergraph convolution in industrial defect detection and use hypergraph computation to process visual features and capture  high-order structures in industrial scenarios. Hyper-YOLO combines typical spatial-domain hypergraph convolution  with  residual connection to propagate high-order messages among vertex feature, performing high-order learning, whose definition is as follows:
\begin{equation}
\left\{
\begin{aligned}
    &x_e = \frac{1}{|\mathcal N_v(e)|} \sum\nolimits_{v \in \mathcal N_v(e)}x_v   {\bm \Theta} \\
    &x_v' = x_v + \frac{1}{|\mathcal N_e(v)|} \sum\nolimits_{e \in \mathcal N_e(v)} x_e
\end{aligned}
\right. ,
\end{equation}
where $\mathcal N_v(e)$ and $\mathcal N_e(v)$ are two neighbor indicate functions, as defined in \cite{HGNNP}: $\mathcal N_v(e) = \{ v \mid v \in e, v \in \mathcal V \}$ and $\mathcal N_e(v) = \{ e \mid v \in e, e \in \mathcal E \}$. ${\bm \Theta}$ is a trainable parameter.  The matrix formulation of  hypergraph convolution is defined as 
\begin{equation}
    \text{HyperConv}(\emph{X}, \emph{H}) = \emph{X} + \emph{D}^{-1}_v \emph{H} \emph{D}^{-1}_e \emph{H}^\top \emph{X} {\bm \Theta},
\end{equation}
where $\emph{D}_v$ and $\emph{D}_e$ represent the diagonal degree matrices of the vertices and hyperedges, respectively. The process of hypergraph construction where incidence matrix $\emph{H}$ is generated is illustrated in Fig. \ref{hgc}.

\subsection{Overview of HD-YOLO}
As shown in Fig. \ref{framework}, we propose a novel industrial defect detection framework named HD-YOLO based on typical YOLO framework. HD-YOLO consists of backbone for extraction, HGANet for aggregation, neck for semantic perception and head for detection.

\begin{figure}
  \centering
  \centerline{\includegraphics[width=0.7\linewidth]{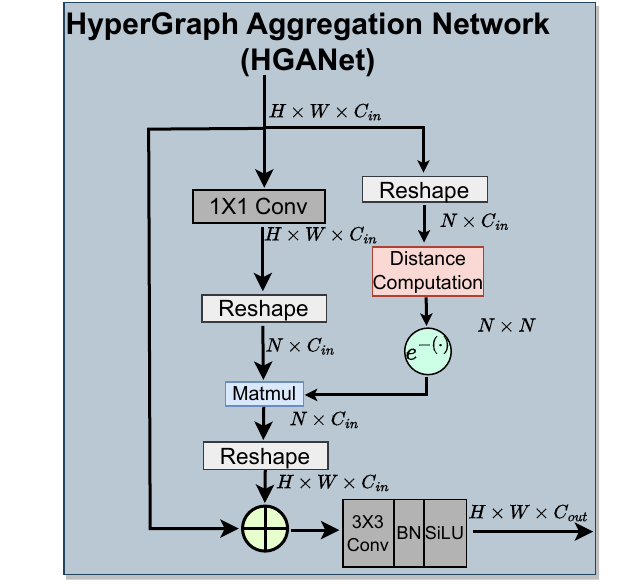}}
  \caption{The structure of HyperGraph Aggregation Network (HGANet).}
  \label{hganet}
\end{figure}

The feature extraction and discernment abilities of backbone are essential for subsequent feature processing. Due to complex background and various sizes of defects, defect features are hard to be extracted unscathed and backbones in YOLO series lack of adaptability for different-scale defects. We replace C3 module found in YOLOv5 with our proposed Defect Aware
Module (DAM). DAM including Defect Aware Network (DAN) can accurately extract key defect features and adapt for uncertain sizes of defects in different industrial scenarios. To enhance exploration of feature interrelationships which is important for defect detection, we combine hypergraph computation with convolution named Mixed Graph Network (MGNet). Hypergraph are used to capture high-order feature interrelationships and convolution are used to explore low-order feature interrelationships. They complement each other in defect feature extraction. 

Multi-scale feature interaction and fusion can dramatically enrich semantic information and facilitate feature extraction and exploration. So before sending features in backbone to neck, we design HyperGraph Aggregation Network (HGANet) to fuse and augment multi-scale features like HyperYOLO. We use negative exponential of distance matrix in hypergraph convolution as attention map and calculate matrix product between attention map and features. HGANet can fuse and enhance multi-scale key defect features and capture more semantic information which is transmitted to neck. 
HGANet plays a pivot bridge between backbone and neck. 

Neck handling both detailed information of shallow features and semantic information of deep features is an important part in YOLO series. Classic YOLO models have limitations in fusion, exploration and facilitation of cross-level features~\cite{hyperyolo}. Cross-Scale Fusion (CSF) consists of channel attention and spatial attention is designed to adaptively fuse cross-level features without loss of information.  Besides, we propose Semantic Aware Module (SAM) to capture enriched semantic features and improve semantic perception of neck. Benefiting from improvement of neck, detection head correspondingly can output accurate detection results.

\begin{figure}
  \centering
  \centerline{\includegraphics[width=0.7\linewidth]{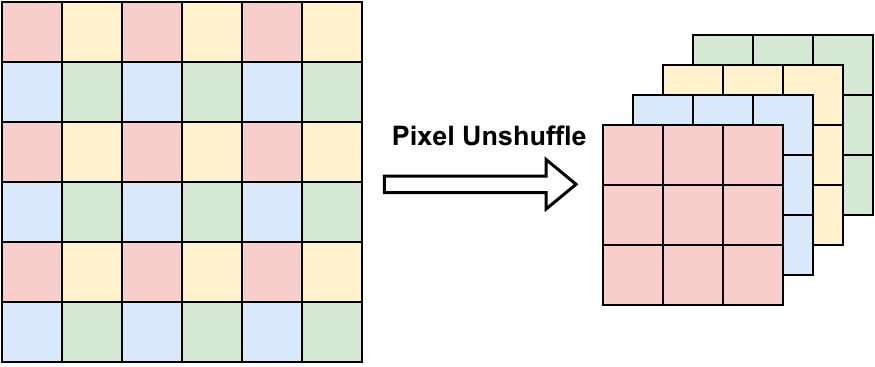}}
  \caption{The scheme of pixel unshuffle.}
  \label{shuffle}
\end{figure}

\subsection{HD-YOLO}
\subsubsection{Defect Aware Module}
To enhance defect feature extraction and mitigate interference caused by sophisticated background in industrial scenarios, we devise the Defect Aware Module (DAM) as shown in Fig. \ref{framework}. DAM use attention mechanism to focus on defect features, mainly consisting of Defect Aware Network (DAN) to generate defect attention map and Conv Block (convolution+BatchNormalization+LeakyReLU) for efficient feature processing and extraction. In DAN, we apply channel attention and spatial attention sequentially on input feature to generate intermediate attention map. Then intermediate attention map and residual connection are operated on feature. The final attention map is sigmoid of intermediate attention map. We use $1 \times 1$ convolution on input feature to generate value embedding in self-attention mechanism. Finally, final attention map and residual connection are imposed on value embedding to generate output of DAM. The formulation of DAN is defined as:
\begin{equation}
    \left\{ 
    \begin{aligned}
        &Avg\_ca = \text{AveragePool}(\emph{X}_{in}) \\ 
        &Max\_ca = \text{MaxPool}(\emph{X}_{in}) \\
        &ca = \text{Sigmoid}(\text{Conv}_1(Avg\_ca \Vert Max\_ca)) \\ 
        &sa = \text{Mean}(\text{Conv}_2(\emph{X}_{in})\cdot ca) \\ 
        &\emph{X}_{mid} = \emph{X}_{in}\cdot sa + \emph{X}_{in} \\ 
        &\emph{X}_{out} = \text{Sigmoid}(\text{Conv}_3(\emph{X}_{mid})) \\
    \end{aligned}
    \right. ,
\end{equation}
where $\Vert$ means concatenation operation.

\subsubsection{Mixed Graph Network}
Convolution blocks in last two stages of original backbone are only for feature downsampling, which is substituted by our proposed Mixed Graph Network (MGNet). MGNet illustrated in Fig. \ref{mgnet} blends primitive convolution and hypergraph convolution in two branches. Feature maps firstly are expanded and then divided averagely along channel dimension. One is sent in hypergraph convolution branch consisting one HyperConv to capture high-order feature interrelationships. While the other part is sent in primitive convolution to explore low-order feature correlations. The outputs from these two branches are concatenated and are processed by $1 \times 1$ bypass convolution for feature recalibration. MGNet is defined as:
\begin{equation}
    \left\{ 
    \begin{aligned}
        &\emph{X}_{ge} = \text{Conv}_1(\emph{X}_{in}) \\ 
        &\emph{X}_{graph}, \emph{X}_{conv} = \text{ChannelSplit}(\emph{X}_{ge}) \\
        &\emph{Y}_{graph} = \text{HyperConv}(\emph{X}_{graph}) \\ 
        &\emph{X}_{le} = \text{Conv}_2(\emph{X}_{conv})\\ 
        &\emph{X}_{conv1}, \emph{X}_{conv2} = \text{ChannelSplit}(\emph{X}_{le}) \\
        &\emph{Y}_{conv2} = \text{BottleNeck}(\emph{X}_{conv2}) \\
        &\emph{X}_{out} = \text{Conv}_4(\emph{Y}_{graph} \Vert \emph{X}_{conv1} \Vert  \emph{Y}_{conv2}) \\
    \end{aligned}
    \right. ,
\end{equation}

\begin{figure}
  \centering
  \centerline{\includegraphics[width=0.7\linewidth]{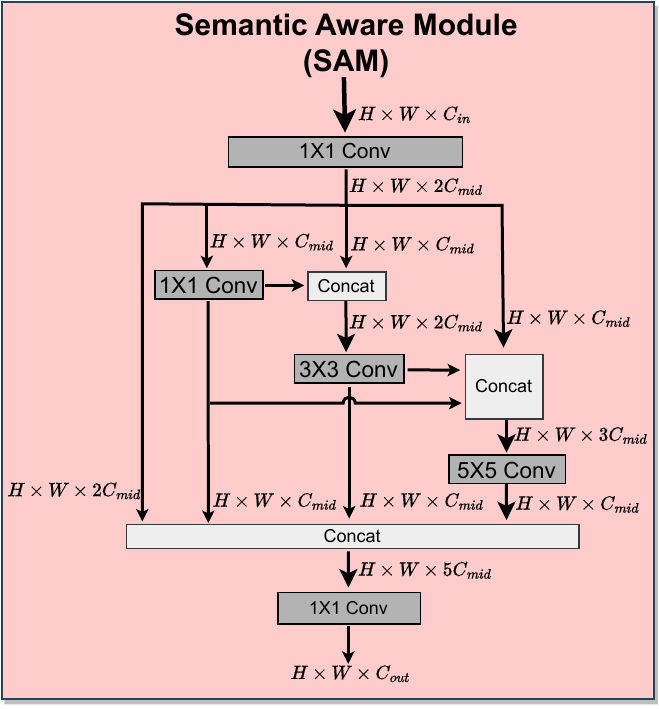}}
  \caption{Illustration of the proposed Semantic Aware Module (SAM).}
  \label{sam}
\end{figure}

\subsubsection{HyperGraph Aggregation Network}
Hyper-YOLO proposes Hypergraph-Based Cross-Level and Cross-Position Representation Network (HyperC2Net) to fuse cross-level and cross-position information from the backbone. By improving HyperC2Net, we propose HyperGraph Aggregation Network (HGANet) to aggregate multi-scale features and explore high-order feature interrelationships. HGANet discards proposed hypergraph convolution in HGNN+ and Hyper-YOLO. After obtaining distance matrix, we calculate negative exponential of distance matrix and treat it as attention map which represents correlations among visual features illustrated in Fig. \ref{hganet}. We dub the attention mechanism based on distance matrix as Distance-Based Attention (DBA). As the distance increases, the value in attention map decreases. As the same way as self-attention mechanism, we use $1 \times 1$ convolution to project input features to value embeddings. Then we calculate matrix product and connect residually with input. HGANet efficiently fuses and augments cross-level and cross-position visual representations. The overarching definition of HGANet shown in Fig. \ref{hganet} is as follows:
\begin{equation}
    \left\{ 
    \begin{aligned}
        &distance = \text{CalculateDistance}(\emph{X}_{in}) \\ 
        &Attention_{dis} = \text{exp}(-distance) \\
        &\emph{V} = \text{Conv}_1(\emph{X}_{in}) \\ 
        &\emph{X}_{mid} = Attention_{dis}\emph{V} + \emph{X}_{in}\\ 
        &\emph{X}_{out} = \text{SiLU}(\text{BN}((\text{Conv}_2(\emph{X}_{mid}))) \\
    \end{aligned}
    \right. ,
\end{equation}

\begin{figure}
  \centering
  \centerline{\includegraphics[width=\linewidth]{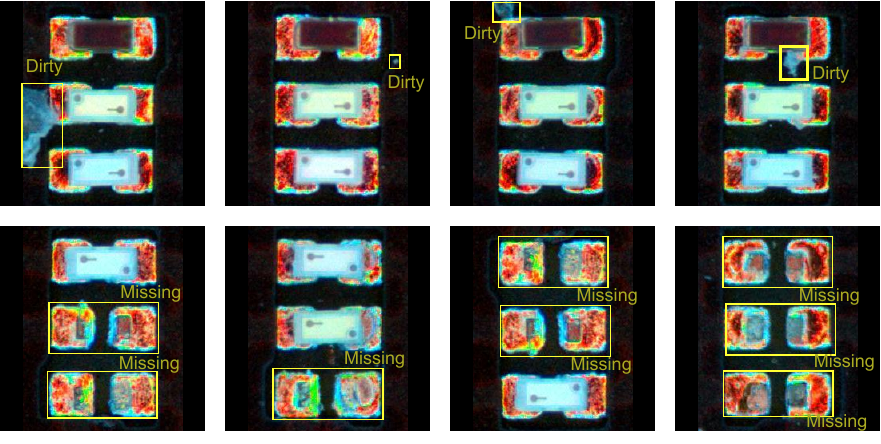}}
  \caption{Examples of defects in MINILED dataset.}
  \label{miniled}
\end{figure}

\subsubsection{Cross-Scale Fusion}
Necks in YOLO series downsample shallow feature maps and concatenate them with deep feature maps, operating $1 \times 1$ convolution for fusion subsequently. However, classical fusion strategy inherently leads to loss of information and is inflexible caused by average pool downsampling and simple concatenation. As a remedy, we propose Cross-Scale Fusion (CSF) indicated in Fig. \ref{framework} in which pixel unshuffle~\cite{rts} is used for downsampling which avoids damage of visual representations and loss of information, especially for tiny defects. The scheme of pixel unshuffle is shown in Fig. \ref{shuffle}. After that, we concatenate two feature maps and use combination of spatial and channel attention to adaptively fuse cross-level features.

\subsubsection{Semantic Aware Module}
To enhance semantic exploration and representation in neck, we propose Semantic Aware Module (SAM) which is a simple expression structure. SAM blends convolutions with different receptive field. SAM engenders diversified information flow and can sufficiently extract multi-scale semantic representation, adapting for various defects with fickle sizes. The architecture of SAM is depicted in Fig. \ref{sam}. It is worth pointing out that we cast off PANet in traditional YOLO neck and only use FPN, reducing the depth and parameters. This is because there are many tiny defects in real-world industrial scenarios. Visual features of tiny defects will disappear as network deepens.

In a word, we propose a series of well-designed modules incorporated in YOLO framework to adapt various sizes of defects and alleviate interference of complex background. In next section, we present thorough experiments to validate workability of these modules.

\begin{figure}
  \centering
  \centerline{\includegraphics[width=\linewidth]{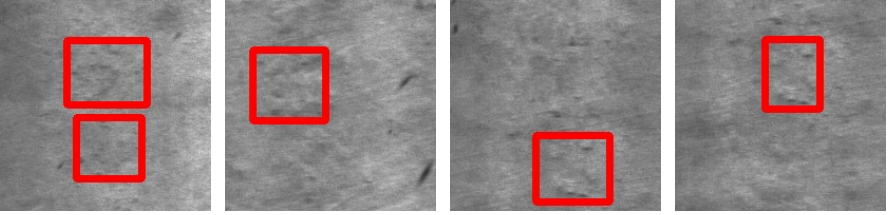}}
  \caption{Hard examples of defects in "Rs" class of NEU-DET dataset.}
  \label{rs}
\end{figure}

\begin{table*}[ht]
\caption{Comparing The Detection Results of Different Methods on HRIPCB dataset}
\label{repcb}
\begin{tabular}{@{}ccccccccccccccc@{}}
\toprule
\multirow{2}{*}{Method} & \multirow{2}{*}{Pre} & \multirow{2}{*}{mAP50} & \multicolumn{2}{c}{Missing\_hole} & \multicolumn{2}{c}{Mouse\_bite} & \multicolumn{2}{c}{Open\_circuit} & \multicolumn{2}{c}{Short} & \multicolumn{2}{c}{Spur} & \multicolumn{2}{c}{Spurious\_copper} \\ \cmidrule(l){4-15} 
& & & mAP50 & Pre & mAP50 & Pre & mAP50 & Pre & mAP50 & Pre & mAP50 & Pre & mAP50 & Pre              \\ \midrule
YOLO-v7~\cite{yolov7}& 86.5& 88.4& 89.6& 84.8& 87.9& 86.5& 87.4& 81.1& 86.2& 88.7& 90.4& 88.2& 89.1& 89.8\\
YOLO-v8~\cite{yolov8}& 87.3& 90.1& 90.1& 85.5& 88.1& 87.3& 88.9& 83.2& 90.9& 89.7& 91.2& 87.6& 91.6& 90.3\\
YOLO-v10~\cite{yolov10}& 89.7& 91.5& 98.9& 94.8& 88.7& 95.0& 96.9& 89.6& 90.1& 82.3& 96.0& 93.3& 78.4& 83.1\\
YOLO-v11~\cite{yolov11}& 96.7& 96.0& 99.5& 98.4& 95.7& 98.0& 99.3& 97.8& 96.4& 90.3& 93.6& \textbf{100}& 91.4& 95.7\\
RT-DETR~\cite{rtdetr}& 87.3& 88.6& 89.1& 85.1& 89.3& 89.7& 86.0& 85.9& 88.4& 89.0& 90.1& 86.5& 88.7& 87.4\\
DetectoRS~\cite{detectors}& 90.4& 93.9& 98.4& 97.3& 97.2& 96.8& 87.9& 86.5& 93.8& 81.0& 91.9& 90.4& 94.1& 90.6\\
PCGAN~\cite{pcgan}& 87.6& 88.7& 88.1& 90.6& 87.7& 86.9& 88.5& 84.6& 88.3& 87.8& 89.2& 85.9& 90.1& 89.9\\
ES-Net~\cite{esnet}& 94.1& 96.3& 97.5& 96.1& 96.3& 96.9& 98.9& 97.3& 95.4& 92.1& 94.7& 88.6& 94.7& 93.6\\
DeYOLOX~\cite{deyolox}& 89.5& 89.3& 88.5& 88.1& 88.3& 89.8& 89.9& 91.3& 88.9& 89.3& 90.3& 90.8& 89.6& 87.9 \\
CA-FPN~\cite{cafpn}& 93.4& 94.7& 96.5& 94.3& 95.6& 96.5& 97.4& 96.3& 95.2 & 92.8& 93.7& 89.1& 89.5& 91.4\\
CCEANN~\cite{cceann}& 93.5& 94.0& 97.8& 95.5& 97.3& 98.0& 96.5& 92.7& 93.6& 92.4& 94.6& 89.6& 84.1& 92.5\\
MSFTI-FDet~\cite{msfti}& 94.7& 96.3& 98.6& 97.1& 96.3& 98.3& 97.0& 96.5& 95.8& 93.0& 95.5& 90.1& 94.8& 93.4\\
DRFA~\cite{drfa}& 96.0& 97.9& \textbf{99.6}& 98.2& \textbf{98.7} & 98.6& 98.5& \textbf{98.3}& 97.6& 94.5& 96.3& 91.6& 96.4& 95.0\\
HyperYOLO~\cite{hyperyolo}& 95.5& 96.1& 99.5& 99.2& \textbf{98.7} & 98.3& 97.9& 97.6& 91.1& 90.7& \textbf{97.6}& 92.7& 91.6& 94.2\\
HD-YOLO (ours)& \textbf{98.1}& \textbf{98.2}& 99.5& \textbf{100}& 98.0& \textbf{100}& \textbf{99.3}& 96.4& \textbf{99.2}& \textbf{97.3}& 94.1& 98.0& \textbf{98.9}& \textbf{97.0}\\ \bottomrule
\end{tabular}
\end{table*}

\begin{table*}[ht]
\caption{Comparing The Detection Results of Different Methods on NEU-DET Dataset}
\label{reneu}
\begin{tabular}{ccccccccccccccc}
\hline
\multirow{2}{*}{Method}              & \multirow{2}{*}{Pre} & \multirow{2}{*}{mAP50} & \multicolumn{2}{c}{Cr} & \multicolumn{2}{c}{Pa} & \multicolumn{2}{c}{Rs} & \multicolumn{2}{c}{Ps} & \multicolumn{2}{c}{Sc} & \multicolumn{2}{c}{In} \\ \cline{4-15} 
& & & mAP50& Pre& mAP50& Pre& mAP50& Pre& mAP50& Pre& mAP50& Pre& mAP50& Pre\\ \hline
YOLO-v7~\cite{yolov7}& 56.6& 74.3& 49.6& 29.3& 83.6& 60.2& 81.9& 61.9& 83.4& 63.8& 59.8& 48.5& 87.7& 75.8\\
YOLO-v8~\cite{yolov8}& 59.6& 77.1& 54.7& 36.7& 85.8& 62.1& 85.3& 67.3& 86.1& 65.3& 61.3& 49.3& 89.4& 76.7\\
YOLO-v10~\cite{yolov10}& 61.5& 64.2& 44.4& 55.0& 73.9& 58.3& 88.2& 72.4& 51.4& 79.8& 49.8& 47.4& 77.8& 56.3\\
YOLO-v11~\cite{yolov11}& 77.5& 80.1& 52.7& 75.5& 87.9& 79.9& \textbf{94.0}& \textbf{82.2}& 86.1& 83.2& 71.4& 70.7& 88.3& 73.3\\
RT-DETR~\cite{rtdetr}& 62.0& 74.7& 50.5& 50.3& 82.9& 66.8& 86.6& 60.3& 79.4& 67.1& 60.1& 50.3& 88.5& 77.3\\
DetectoRS~\cite{detectors}& 61.2& 80.3& 58.9& 42.3& 88.3& 61.2& 88.5& 69.9& 87.9& 66.9& 62.7& 50.7&\textbf{95.7}& 76.3\\
PCGAN~\cite{pcgan}& 56.3& 77.1& 53.6& 37.7& 85.6& 54.2& 84.7& 63.9& 86.5& 64.6& 62.1& 45.6& 90.1& 71.9\\
ES-Net~\cite{esnet}& 55.3& 79.1& 56.0& 28.8& 87.6& 55.6& 88.3& 68.2& 87.4& 65.8& 60.4& 43.1& 94.9& 70.5\\
DeYOLOX~\cite{deyolox}& 57.0& 76.6& 52.1& 38.6& 87.3& 56.5& 86.3& 64.5& 85.0& 65.9& 60.1& 43.2& 88.9& 73.2\\
CA-FPN~\cite{cafpn}& 60.6& 79.5& 58.3& 43.6& 87.6& 56.9& 86.7& 70.1& 88.3& 67.3& 61.9& 50.1& 94.2& 75.4\\
CCEANN~\cite{cceann}& 58.7& 78.5& 53.9& 40.1& 88.4& 59.1& 88.7& 67.6& 87.6& 66.5& 60.9& 44.0& 91.5& 74.6\\
MSFTI-FDet~\cite{msfti}& 58.7& 79.0& 55.4& 39.4& 88.9& 58.7& 89.3& 65.2& 86.3& 67.1& 61.4& 45.9& 92.6& 75.9\\
HyperYOLO~\cite{hyperyolo}& 76.4& 80.7& 57.2& 66.5& 88.6& 79.6& 93.8& 84.1& 87.8& 90.2& 64.9& 57.5& 92.1& \textbf{80.1} \\
\multicolumn{1}{l}{HD-YOLO (ours)} & \textbf{78.8}& \textbf{81.6}& \textbf{68.9}& \textbf{83.3}& \textbf{92.9}& \textbf{82.6}& 58.3& 57.5& \textbf{88.7}& \textbf{90.1}& \textbf{92.}9& \textbf{80.4}& 87.8& 79.1\\ \hline
\end{tabular}
\end{table*}

\begin{table}[ht]
\caption{Comparing The Detection Results of Different Methods on MINILED  Dataset}
\label{reminiled}
\resizebox{\linewidth}{!}{
\begin{tabular}{ccccccc}
\hline
\multirow{2}{*}{Method} & \multirow{2}{*}{Pre} & \multirow{2}{*}{mAP50} & \multicolumn{2}{c}{dirty} & \multicolumn{2}{c}{missing} \\ \cline{4-7} 
& & & mAP50& Pre& mAP50& Pre\\ \hline
Faster-RCNN~\cite{fasterrcnn}& 80.6& 89.8& 86.8& 66.3& 92.7& 95.0\\
Cascade-RCNN~\cite{cascadercnn}& 70.6& 88.5& 86.9& 66.1& 90.2& 75.0\\
Libra-RCNN~\cite{librarcnn}& 80.0& 93.5& 87.0& 60.0& \textbf{100}& \textbf{100}\\
RetinaNet~\cite{retinanet}& 73.0& 93.1& 86.3& 63.7& \textbf{100}& 82.4\\
FCOS~\cite{fcos}& 92.4& 94.2& 88.5& 84.7& \textbf{100}& \textbf{100}\\
YOLOv8~\cite{yolov8}& \textbf{93.0} & 94.4& 89.2& 87.9& 99.5& 98.2\\
YOLOv10~\cite{yolov10}& 89.9& 93.1& 87.3& 87.0& 98.9& 92.7\\
YOLOv11~\cite{yolov11}& 92.3& 93.6& 87.7& 87.4& 99.5& 97.1\\
Hyper-YOLO~\cite{hyperyolo}& 91.7& 94.2& 65.7& \textbf{88.9}& 92.4& 99.5\\
HD-YOLO (ours)& 92.5& \textbf{94.9}& \textbf{90.3}& 86.5& 99.5& 98.6\\ \hline
\end{tabular}}
\end{table}

\section{Experiments}
\subsection{Dataset Description}
To evaluate the effectiveness of our proposed HD-YOLO, we conducted extensive experiments on three different industrial datasets. HRIPCB and NEU-DET are public datasets. MINILED is self-built dataset collected by our Automated Optical Inspection (AOI) machine. The details of the datasets are as follows:
\par 1) HRIPCB: It is a printed circuit board dataset collected by Peking University, which contains 693 pcb images with average  $2777 \times 2188$ high resolution. There are six types of defects in HRIPCB, that is missing hole,  open circuit, spur, short, spurious copper and mouse bite. The division ratio of the training set and the test set in our experiment is 9:1.

\par 2) NEU-DET: It is one of the most commonly used industrial dataset about hot-rolled steel plates. There are also six types of defects in NEU-DET including crazing, inclusion, patches, pitted surface, rolled-in scales, and scratches.  Each defect has 300 images.

\par 3) MINILED: MINILED dataset is self-built with our Automated Optical Inspection machine. It has 2050 miniled images with two defect classes consisting of dirty and missing which are called "defect" and "lg" in annotations shown in Fig. \ref{miniled}.

\begin{table*}[h]
\centering
\caption{Ablation Study on Effectiveness of Proposed Modules}
\label{pm}
\begin{tabular}{cccccccccccc}
\hline
\multirow{2}{*}{baseline} & \multirow{2}{*}{DAM} & \multicolumn{1}{l}{\multirow{2}{*}{HGANet}} & \multirow{2}{*}{SAM} & \multirow{2}{*}{MGNet} & \multirow{2}{*}{CSF} & \multicolumn{2}{c}{NEU-DET}  & \multicolumn{2}{c}{PCB} & \multicolumn{2}{c}{MINILED}\\ \cline{7-12} 
& & \multicolumn{1}{l}{} & & & & mAP50& \multicolumn{1}{l}{Pre} & mAP50 & \multicolumn{1}{l}{Pre} & mAP50 & \multicolumn{1}{l}{Pre}\\ \hline
\checkmark& & & & & & 74.3& 56.6&87.7 &94.3 &93.6 &89.2\\
& \checkmark& \checkmark& & & & 78.3  & 71.4 & 96.9  & 97.2 & 94.3 & 91.7\\
& \checkmark& \checkmark& \checkmark& & & 79.1  & 76.0& 98.0& 97.6 & 94.1 & 91.1\\
& \checkmark& \checkmark& \checkmark& \checkmark& & 79.2  & 77.4& 97.6& 95.2 & 94.2 & 91.8\\
& \checkmark& \checkmark& \checkmark& \checkmark& \checkmark   & 81.6  & 78.8& 98.2& 98.1 & 94.9 & 92.5\\
& \checkmark& \checkmark& \checkmark&  & \checkmark   & 78.5  & 78.5& 97.7& 97.5 & 94.2 & 91.8\\\hline
\end{tabular}
\end{table*}

\subsection{Experimental Setup}
\subsubsection{Evaluation Metrics}
We use common detection metrics including precision ($Pre=\frac{TP}{TP+FP}$), average precision ($AP=\int_{0}^{1}PdR$) and mean average precision ($mAP=\frac{1}{n+1}\sum_{i=1}^NAP_i$) as our performance indicators.
 
\subsubsection{Implementation Details}
The implementation is based on the Ultralytics library on Nvidia 3090 GPU. 
In the experiment, the size of the input image is set to $1088 \times 1088 $ for HRIPCB,  $640 \times 640$ for NEU-DET and MINILED. Limited by hardware, batch size is 2 for HRIPCB and 16 for  NEU-DET and MINILED. We select SGD as optimizer with initial learning rate of 0.01 and a weight decay of 0.0005. Momentum is set to 0.937. Epoch is set to 500. Hypergraph threshold is set to 3, 6, 6 for HRIPCB, NEU-DET and MINILED.

\begin{table}[ht]
\caption{Ablation Study on Threshold of Hypergraph Construction.}
\label{th}
\begin{tabular}{ccccccc}
\hline
\multirow{2}{*}{Threshold}  & \multicolumn{2}{c}{NEU-DET} & \multicolumn{2}{c}{HRIPCB} & \multicolumn{2}{c}{MINILED} \\ \cline{2-7} 
 & Pre& mAP50& Pre& mAP50& Pre& mAP50\\ \hline
\multicolumn{1}{c}{3}& 73.3& 79.3& 98.1& 98.2& 91.4& 94.9\\ 
\multicolumn{1}{c}{4}& 73.0& 78.3& 97.0& 98.0& 92.3& 94.0\\
\multicolumn{1}{c}{5}& 72.8& 79.8& 97.1& 97.6& 92.3& 94.7\\
\multicolumn{1}{c}{6}& 78.8& 81.6& 98.6& 97.9& 92.5& 94.9\\
\multicolumn{1}{c}{8}& 76.6& 80.1& 98.6& 97.9& 91.3& 94.9\\
\hline
\end{tabular}
\end{table}

\begin{table}[ht]
\caption{Ablation Study on Receptive Field in SAM Module.}
\label{rf}
\begin{tabular}{ccccccc}
\hline
\multirow{2}{*}{Receptive Field}  & \multicolumn{2}{c}{NEU-DET} & \multicolumn{2}{c}{HRIPCB} & \multicolumn{2}{c}{MINILED} \\ \cline{2-7} 
 & Pre& mAP50& Pre& mAP50& Pre& mAP50\\ \hline
\multicolumn{1}{c}{\{1, 3, 5\}}& 78.8& 81.6& 98.4& 98.2& 92.5& 94.9\\ 
\multicolumn{1}{c}{\{3, 3, 3\}}& 73.4& 79.6& 96.6& 97.5& 90.8& 94.5\\
\multicolumn{1}{c}{\{1, 5, 7\}}& 72.9& 78.2& 96.7& 97.9& 90.4& 94.4\\
\multicolumn{1}{c}{\{3, 5, 7\}}& 74.0& 80.4& 95.2& 97.3& 90.1& 94.6\\
\multicolumn{1}{c}{\{5, 5, 5\}}& 80.7& 79.6& 97.7& 97.8& 91.5& 94.7\\
\hline
\end{tabular}
\end{table}

\subsection{Results and Discussions}
We conduct comprehensive comparison between our proposed HD-YOLO and other state-of-the-art models on three datasets including classical object detection methods and task-specific detection methods. Task-specific detection methods refer to those detection methods specializing for industrial defect detection, such as DRFA~\cite{drfa}, ES-Net~\cite{esnet}, CANet~\cite{canet}, etc. 

Firstly, the performance of HD-YOLO on HRIPCB dataset is illustrated in Table \ref{repcb}. It is demonstrated that our proposed HD-YOLO outperforms all other detection methods on two commonly used metrics $Pre$ and $mAP_{0.5}$. To be specific, precision $Pre$ of HD-YOLO achieves 98.1\% and $mAP_{0.5}$ reaches 98.2\%. Compared with state-of-the-art task-specific detection method DRFA, HD-YOLO achieves an improvement of 2.1 and 0.3 in terms of $Pre$ and $mAP_{0.5}$. Even if compared with latest general detection methods YOLO-v10 and YOLO-v11, HD-YOLO shows an improvement of (8.4\%, 6.7\%) and (1.4\%, 2.2\%) about ($Pre$, $mAP_{0.5}$) respectively. Considering that HD-YOLO incorporates hypergraph computation in YOLO framework inspired by HyperYOLO, we note that HD-YOLO outperforms HyperYOLO by 2.6\% and 2.1\% on $Pre$ and $\text{mAP}_{0.5}$. Compared with ES-Net specializing for tiny defects detection, HD-YOLO still reaches a gain of 4.0\% in $Pre$ and 1.9\% in $\text{mAP}_{0.5}$.  It is noteworthy that our method achieves 100\% $Pre$ on class "Missing hole" and "Mouse bite" in HRIPCB dataset. These results validate the effectiveness of the HD-YOLO method on tiny defects detection.

As for the performance on larger defects, we use NEU-DET dataset to verify generalization and effectiveness of HD-YOLO. As showcased in Table \ref{reneu}, we found that the proposed HD-YOLO achieves the highest object detection performance (78.8\%, 81.6\%) in terms of ($Pre$, $\text{mAP}_{0.5}$), improving state-of-the-art methods Hyper-YOLO by 2.4\% and 0.9\% respectively. HD-YOLO outperforms latest general detection methods YOLO-v10 and YOLO-v11 by (17.3\%, 17.4\%) and (1.3\%, 1.5\%). It is noticeable that HD-YOLO demonstrates significant performance improvements on "Cr" (crazing) and "Sc" (scratches) class of NEU-DET, achieving more than 10\% in both $Pre$ and $\text{mAP}_{0.5}$ compared with other methods. When compared to transformer-based detection method RT-DETR, the improvements brought by our HD-YOLO are significant (from 62.0 to 78.8 on $Pre$ and from 74.7 to 81.6 on $\text{mAP}_{0.5}$). However, it can be found that HD-YOLO engenders drastic performance drop on "Rs" (rolled-in scales) class. After analyzing, we find that defects in "Rs" class are very similar with background and difficult to detect depicted in Fig 7, causing terrible performance.  To sum up, HD-YOLO can be applied in different industrial scenarios and achieve a satisfactory results in larger defect detection.

\begin{figure}[ht]
  \centering
  \centerline{\includegraphics[width=\linewidth]{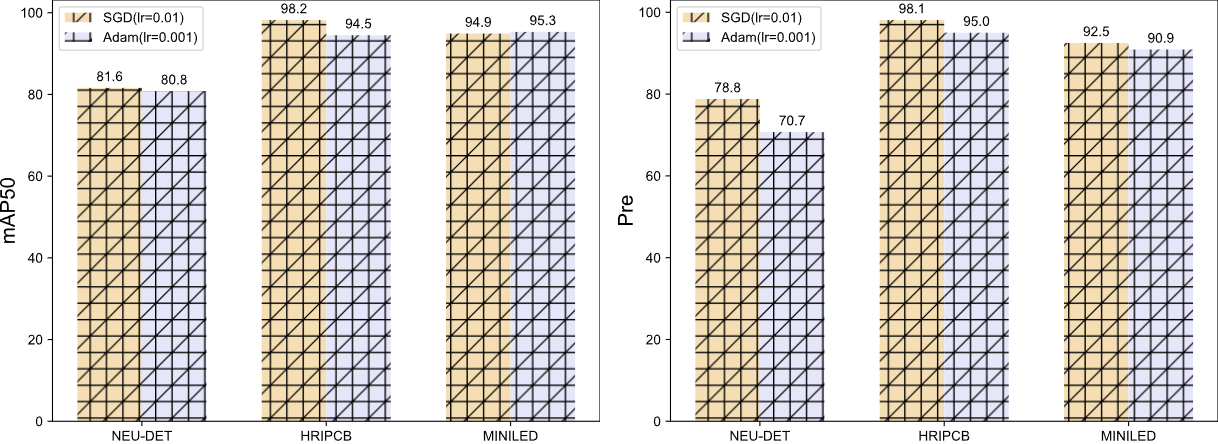}}
  \caption{Comparison of the SGD and Adam optimizers for HD-YOLO.}
  \label{opt}
\end{figure}

Additionally, we apply HD-YOLO in our real-world industrial scenarios where dirty and missing defects in miniled are needed to be detected. The performance of HD-YOLO on MINILED dataset is shown in Table \ref{reminiled}. Our proposed HD-YOLO achieves state-of-the-art performance in 94.9\% $\text{mAP}_{0.5}$ and improves latest method YOLO-v11 by 1.3\%. When compared to Hyper-YOLO, improvements on $Pre$ and $\text{mAP}_{0.5}$ are 0.8\% and 0.7\% respectively. Except for YOLO series, we also make comparison with two-stage detection methods RCNN series including Faster-RCNN, Cascade-RCNN and Libra-RCNN. The improvements on $Pre$ are more than 10\% compared to RCNN series. HD-YOLO outperforms Faster-RCNN by 5.1\% $\text{mAP}_{0.5}$ and outperforms Libra-RCNN by 1.4\%. Performing well on MINILED dataset demonstrates that HD-YOLO is generalized and practical.

\begin{figure*}
\centering
\includegraphics[width=0.9\textwidth,]{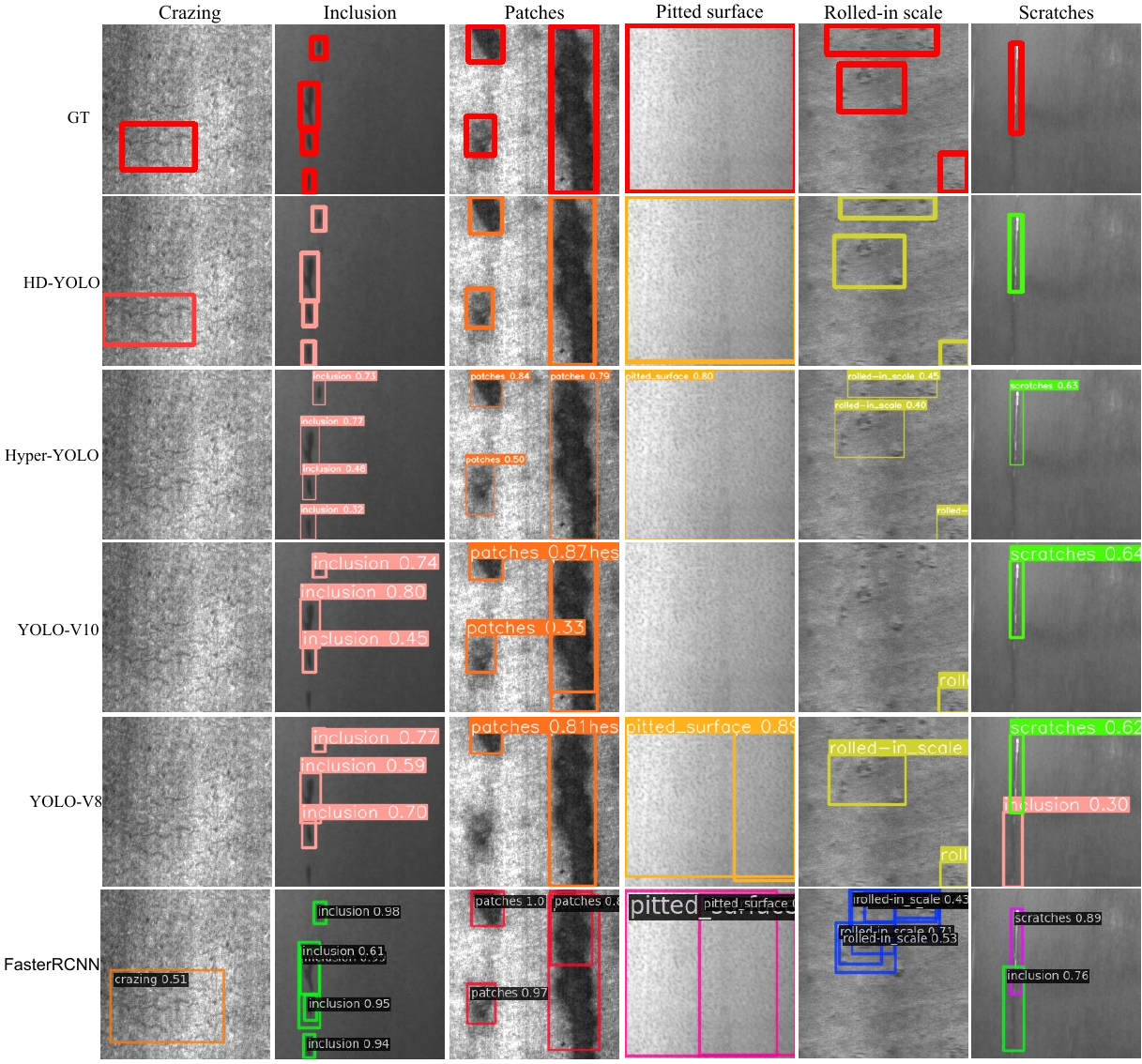}\\
\caption{Results comparison of HD-YOLO and other detection methods on NEU-DET dataset. }
\label{visneu}
\end{figure*}

\subsection{Effectiveness of Proposed Modules}
To validate effectiveness of our proposed modules in HD-YOLO, we conduct a series of experiments in which modules are incorporated in baseline step by step illustrated in Table \ref{pm}. We regard YOLO-v5 framework as our baseline. To begin with, we add DAM and HGANet into YOLO framework. DAM specializes for defect detection and HGANet combines hypergraph and attention mechanism to aggregate multi-scale features. The performance is improved by (14.8\%, 4.0\% ), (2.9\%, 9.2\%) and (2.5\%, 0.7\%) with respect to ($Pre$, $\text{mAP}_{0.5}$) on NEU-DET, HRIPCB and MINILED respectively. After replacing C3 block of YOLO-v5 with our proposed SAM in neck,  $\text{mAP}_{0.5}$ is improved by 0.8\%, 1.1\% on NEU-DET and HRIPCB. SAM can effectively explore semantic visual features. MGNet incorporates hypergraph computation in backbone and aims to capture high-order feature interrelationships. MGNet leads to 0.1\% and 1.4\% improvements on NEU-DET and 0.1\% and 0.7\% improvements on MINILED in terms of $\text{mAP}_{0.5}$ and $Pre$ but suffers from performance deterioration on HRIPCB. This is because defects in HRIPCB are so tiny that high-order feature interrelationships are weak and hard to capture. Finally, we use CSF to substitute for downsampling and concatenation in the neck where we adopt pixel unshuffle for downsampling without loss of information and combination of channel and spatial attention mechanism to fuse cross-scale features.  With CSF, $\text{mAP}_{0.5}$ achieve 81.6\%, 98.2\% and $Pre$ achieve 78.8\% and 98.1\% on NEU-DET, HRIPCB, which are state-of-the-art performance. Besides, HD-YOLO achieves the best performance on our self-built MINILED datasets to validate the generalization of our proposed modules. Due to performance degradation caused by MGNet, we remove MGNet from HD-YOLO framework to verify the effectiveness of MGNet specifically. As can be seen from the last row in Table \ref{pm}, both $\text{mAP}_{0.5}$ and $Pre$ on three datasets show a certain level of decrease.

\begin{table*}[h]
\caption{Comparison of Operating Efficiency of Different Methods (Dataset: NEU-DET, Training, Inference Size: 640, And Epochs: 300)}
\label{oe}
\resizebox{\linewidth}{!}{
\begin{tabular}{ccccccccccccc}
\hline
Method & YOLO-v7 & YOLO-v8 & RT-DETR & DetectoRS & PCGAN  & ES-Net & DeYOLOX & CAFPN  & CCEANN & MSFTI-FDet & DRFA & Ours \\ \hline
Inference time & 19.5ms& 18.1ms& 20.3ms& 78ms& 22.4ms& 35.2ms& 16.5ms& 70.2ms& 140ms& 18.2ms& 12.1ms&   5.4ms   \\
Training time  & 4.1h& 4.0h& 4.4h& 4.8h& 3.6h& 3.5h& 3.1h& 3.8h& 4.0h& 3.3h& 3.2h&   1.4h   \\
Parameters     & 71.9M& 68.2M& 67.4M& 80.4M& 80.4M& 148M& 12.3M& 36.8M& 167M& 22.5M& 11.8M&   9.3M   \\ \hline
\end{tabular}}
\end{table*}

\begin{figure*}
\centering
\includegraphics[width=\textwidth,]{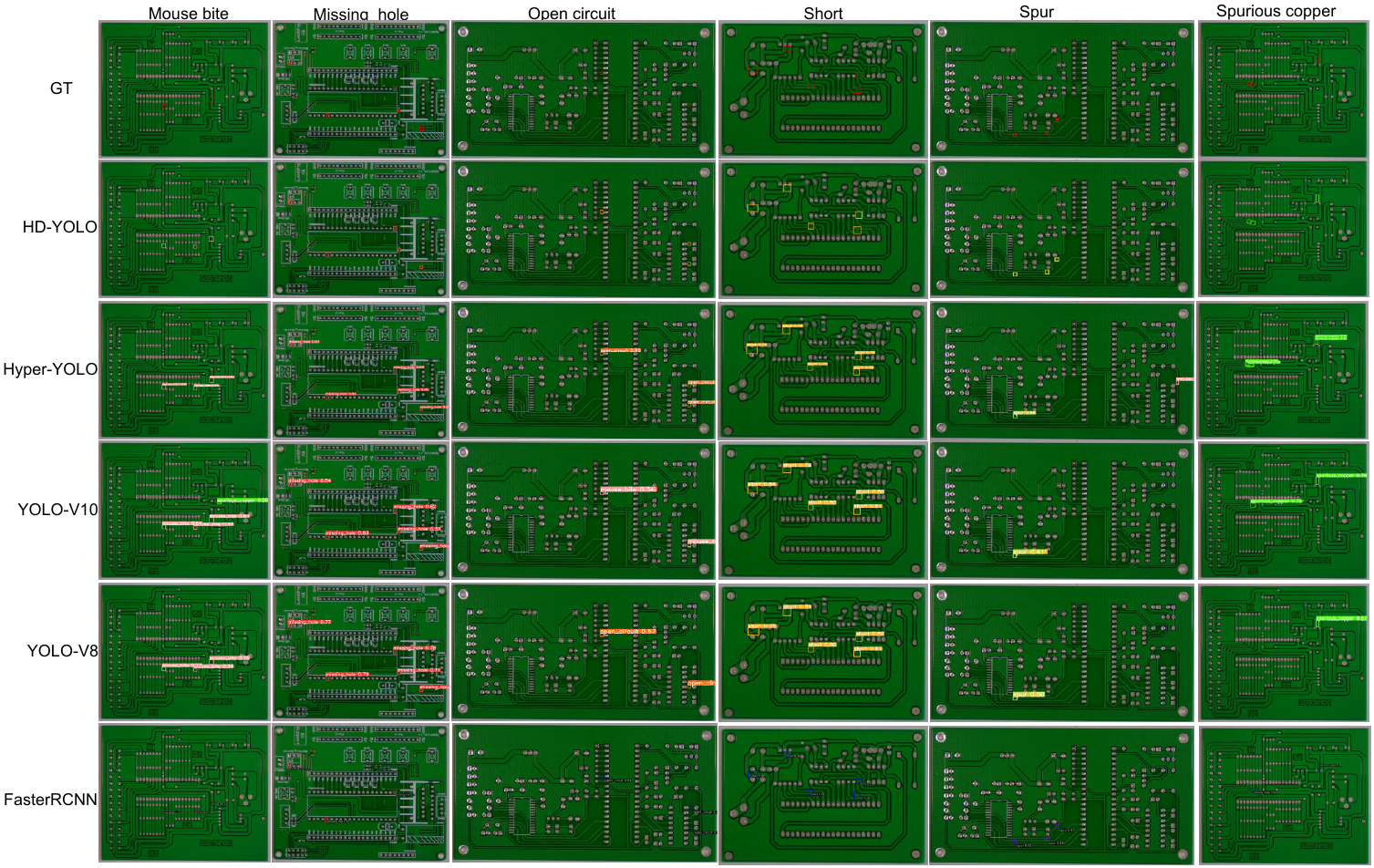}\\
\caption{Results comparison of HD-YOLO and other detection methods on HRIPCB dataset. }
\label{vispcb}
\end{figure*}

\begin{figure*}
\centering
\includegraphics[width=0.9\textwidth,]{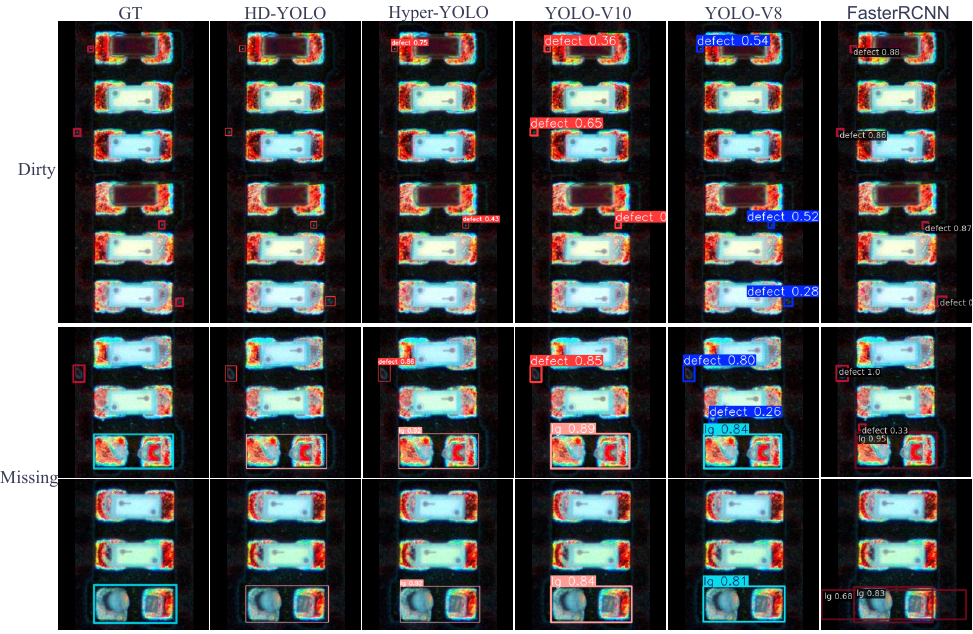}\\
\caption{Results comparison of HD-YOLO and other detection methods on MINILED dataset. }
\label{visminiled}
\end{figure*}

\subsection{Ablation Studies}
\subsubsection{Influence of hypergraph threshold}
Hypergraph threshold $\epsilon$, involving in hypergraph construction, plays a key role in hypergraph convolution. Hypergraph threshold $\epsilon$ controls the number of nodes and edges in hypergraph. As mentioned in HyperYOLO, larger threshold will lead to dense-connected hypergraph and lower threshold will lead to sparse-connected hypergraph. In this section, we will figure out the influence of threshold $\epsilon$ in defect detection. We conduct a series of experiments on three datasets with different hypergraph threshold $\epsilon$.  As shown in Table \ref{th}, we find that $\text{Pre}$ and $\text{mAP}_{0.5}$ of HD-YOLO on HRIPCB achieve the best when $\epsilon$ is set to 3. Though $Pre$ are higher when $\epsilon$ is 6, we choose $\text{mAP}_{0.5}$ as our main performance indicator. With regard to NEU-DET and MINILED, $\epsilon=6$ is the optimal hypergraph threshold observed from Table \ref{th}.  We experimentally demonstrate that too small threshold will hurdle exploration of feature interrelationships and too large threshold will lead to more redundant representations and interrelationships which results in performance drop. The reason why small threshold is suitable in HRIPCB is the the defects in HRIPCB are so tiny that visual feature interrelationships are restricted.

\subsubsection{Effectiveness of Receptive Field in SAM}
In our proposed Semantic Aware Module (SAM), convolutions with different kernel sizes endow this module multi-receptive field awareness ability. How to choose optimal kernel size is essential for better exploring semantic visual features.
We set kernel sizes to \{1, 3, 5\} in SAM. To validate the effectiveness of our chosen kernel sizes, we change the kernel sizes and conduct more experiments. The results are shown in Table \ref{rf}. It can be found that the results are best on our chosen kernel sizes \{1, 3, 5\}. When we set kernel sizes to 3 rigidly or to \{1, 5, 7\}, \text{Pre} and $\text{mAP}_{0.5}$ both degrade a lot. The precision \text{Pre} drops from 78.8\% to 72.9\% and $\text{mAP}_{0.5}$ drops from 81.6\% to 78.2\% on NEU-DET dataset. As for MINILED and HRIPCB, the precision \text{Pre} decreases in the 1$\sim$2\% range. Even if we increase the whole kernel sizes to  \{3, 5, 7\} as shown in fourth row of Table \ref{rf}, there are no improvements on \text{Pre} and $\text{mAP}_{0.5}$ compared to kernel sizes \{1, 3, 5\} but compared to \{3, 3, 3\} and \{1, 5, 7\} \text{Pre} and $\text{mAP}_{0.5}$ are improved a little on NEU-DET and MINILED dataset. This is because larger kernel sizes can extract more information of larger defects in NEU-DET. When kernel sizes are set to \{5, 5, 5\}, results on NEU-DET, HRIPCB and MINILED are improved and more close to the performance of HD-YOLO. The precision \text{Pre} on NEU-DET dataset even surpasses HD-YOLO. However, large kernel sizes lead to more parameters. Kernel sizes \{1, 3, 5\} are optimal considering no matter performance or efficiency.

\subsubsection{Different Optimizers for HD-YOLO}
We use Adam and SGD optimizers to train HD-YOLO on three datasets mentioned above and compare the performance under different optimizers. $\text{mAP}_{0.5}$ and \text{Pre} are presented in Fig. \ref{opt} when using SGD and Adam. It demonstrates that though Adam is an useful optimizer widely used in other methods, it is suboptimal when compared to SGD in HD-YOLO. The $\text{mAP}_{0.5}$ and \text{Pre} drop when using Adam on HRIPCB and NEU-DET. \text{Pre} on MINILED is improved by 0.4\% using Adam optimizer. Notably, learning rate need to be set to 0.001 for Adam and 0.01 for SGD. The performance is so terrible when learning rate is 0.01 for Adam that we do not post the results in this paper. Considering the whole performance on three datasets, we use SGD optimizer to train HD-YOLO.

\subsubsection{Efficiency of HD-YOLO}
Though our proposed HD-YOLO demonstrates a satisfactory performance on different industrial datasets, efficiency is also a key factor in real-world application. We make comprehensive evaluation and comparison on NEU-DET dataset in terms of efficiency depicted by inferencing time, training time and number of parameters. It can be seen from Table \ref{oe} that HD-YOLO costs the least inferencing time, down to 5.4 ms (185.1 FPS),  which is more than twice as fast as the most efficient method DRFA. The speed of HD-YOLO largely meets the need of industrial practical application. With 300 training epochs, HD-YOLO just needs 1.4 h, which is far less time-costly than other methods. HD-YOLO can be transferred quickly in different scenarios attributing to less training time. The number of parameters is a dispensable indicator to measure the efficiency of models. The number of parameters of HD-YOLO is 9.3 M, which is light-weight compared with other detection methods. In a nutshell, HD-YOLO exceeds other state-of-the-art methods on operational efficiency with respect to inferencing time, training time and number of parameters with a large margin.

\subsection{Visualization of HD-YOLO in Industrial Defect Detection}
For a qualitative comparison, we visualize the detection results on three datasets of HD-YOLO and other state-of-the-art methods. We provide visualization of the detection results on NEU-DET dataset in Fig. \ref{visneu}. It can be found that detection results of HD-YOLO are more accurate than other methods. Defects are very similar with background in some classes of NEU-DET like "Crazing", "Pitted surface" and "Rolled-in scale". So some not apparent defects are missing in detection results of other methods as shown in first and fifth columns in Fig. \ref{visneu}. However, HD-YOLO can focus on defect features and weaken the interference of background, leading to more accurate detection results. The visualization of detection results on HRIPCB are shown in Fig. \ref{vispcb}. As can be observed, in HRIPCB, the sizes of defects are universally tiny and the background are complicated which raises the difficulty of detection. Therefore, we found that the detection results of YOLO-v10, YOLO-v5 and YOLO-v11 are inaccurate and they can not detect all defects in pcb images and even give incorrect defects classification demonstrated in results of Hyper-YOLO. While HD-YOLO can effectively extract visual features of tiny defects and achieves the best performance against all other methods in six classes of HRIPCB. For practical application, we additionally present the detection results of HD-YOLO and some detection methods on MINILED dataset in Fig. \ref{visminiled}. As can be seen, our HD-YOLO still has the best performance for defect detection which are closest to the ground truth. Especially in tiny defects, HD-YOLO hardly miss tiny defects while uncaught defects appear in other methods.  

In a word, the visualization of detection results demonstrates the strong detection ability and generalization of our HD-YOLO across diverse industrial scenarios. Compared with existing methods, HD-YOLO outstands by three key points: 1) effective extraction and exploration of visual features of tiny defects, 2) robustness to intricate background, and 3) generalization across different products in industrial scenarios.

\section{Conclusion}
In this paper, we presented a novel industrial defect detection framework HD-YOLO. In HD-YOLO, we integrate hypergraph computation in YOLO framework and design a series visual modules specializing for defect detection. For effective extraction of tiny defect features and capture high-order feature interrelationships in backbone, we propose Defect Aware Module (DAM) and Mixed Graph Network (MGNet) to improve visual exploration ability. Inspired by HyperYOLO, we propose HyperGraph Aggregation Network (HGANet) to aggregate multi-scale features using our proposed distance-based attention before neck.
To enhance semantic perception and adaptive fusion ability of neck, we propose CSF (Cross-Scale Fusion) and Semantic Aware Module (SAM) modules as a remedy.
Thorough experiments on both NEU-DET and HRIPCB benchmark illustrate the effectiveness of all modules and superior performance in industrial defect detection of HD-YOLO. Besides, we evaluate HD-YOLO on our self-built MINILED dataset for practical application. The results also show that HD-YOLO gives a feasible and practical solution for accurately detecting various defects in industrial scenarios.

\bibliographystyle{IEEEtran}
\bibliography{IEEEabrv,reference}


\end{document}